\documentclass[shortAfour,times]{sagej}

\usepackage[natbibapa]{apacite}
\usepackage{verbatim}
\usepackage{todonotes}
\usepackage{lipsum}
\usepackage[export]{adjustbox}
\usepackage{subfig}
\usepackage[hidelinks]{hyperref}

\definecolor{cambridgeblue}{rgb}{0.64, 0.76, 0.68}

\def\volumeyear{2022}
\def\journalname{Adaptive Behavior}

\newenvironment{junk}
{ \comment }
{ \endcomment  }

\newcommand{\icol}[1]{% inline column vector
  \left(\begin{smallmatrix}#1\end{smallmatrix}\right)%
}

\urldef{\jhsgithub}\url{https://github.com/jstovold/CogSis_PythonSim/}

\newcommand{\ie}{i.\,e.~}

\newcommand{\eg}{e.\,g.~}

\newcommand{\nhypothesis}{\paragraph{Null Hypothesis ($H_0$):}}
\newcommand{\param}[1]{\texttt{#1}}

\newcommand{\cites}[1] {\citeauthor{#1}'s~(\citeyear{#1})}

\newcommand{\cah}{CogSis} % cognitive adaptive homeostasis

\newcommand{\cahLongUpper} {COGnitive HomeostaSIS}

\newcommand{\cdm}{CDM} 
\newcommand{\cdmLong} {Cognitive Decision Making}
\newcommand{\cdmLongLower}{cognitive decision making}

\newcommand{\ii}{II} % cognitive adaptive homeostasis
\newcommand{\iiLong} {Internal Image}
\newcommand{\iiLongLower} {internal image}

\newcommand{\resQu}[1]{RQ#1}

\begin{document}
\runninghead{James Stovold et al.}

\title{Cognition-inspired homeostasis can balance conflicting needs in robots}

\author{James Stovold\affilnum{1}, Simon O'Keefe\affilnum{2}, and Jon Timmis\affilnum{3}}

\affiliation{\affilnum{1} School of Computing and Communications, Lancaster University in Leipzig \\ 
\affilnum{2} York Cross-Disciplinary Centre for Systems Analysis, University of York \\
\affilnum{3} School of Computer Science, University of Sunderland}

\corrauth{J.\ H.\ Stovold, Lancaster University in Leipzig,
Nikolaistra\ss{}e 10, 04109 Leipzig, Germany}

\email{j.stovold@lancaster.ac.uk}

\begin{abstract}

 Homeostasis keeps animals alive; it is a fundamental process that allows animals to adapt quickly to their 
 environment. Artificial homeostasis can be used to help robots adapt to changing environments. Previous attempts at 
 developing artificial homeostasis for robots were driven by mimicry of the biochemical machinery that drives 
 homeostasis in humans. By considering homeostasis from a cognitive perspective, we develop a comparatively simple 
 robot controller named \cah{} (\cahLongUpper{}) and demonstrate that it can provide homeostasis to a robot, even 
 when there are conflicting needs. We present experiments showing that a robot running \cah{} is able to learn from 
 previous experiences and use them to influence future behaviour; can maintain its charge level while attending to 
 another task (warming itself in an area separate from the charging station); and is able to maintain its charge 
 level while avoiding a conflicting need (keeping cool, when the charging station is placed in a hot region of the 
 environment). Results are presented in simulation and from a real robot platform.
\newline
\newline
 Code for simulations and the robot controller are available at: \newline \jhsgithub{}
\end{abstract}


\maketitle
%\newpage{}

\section{Introduction}
\label{sec:introduction}

\begin{junk}

by altering our perception of the world around us, the processes of homeostasis changedare able to encourage certain 
conscious actions and discourage others, thereby maintaining the body's internal environment
the processes of homeostasis are able to encourage 
certain conscious actions and discourage others, thereby maintaining the body's internal environment.

\end{junk}

 Homeostasis is a fundamental process in living beings, allowing our bodies to adapt to changes in the environment. 
The mechanisms to maintain the biochemical environment of the body have been studied at great length since 1878 when 
Claude Bernard observed that the internal environment (or `{\em milieu interne}') was essential to 
survival~\citep{bernard_lesphenomenesdelavie, cannon_organizationhomeostasis}.

\citet{neal_timidityusefulemotional,vargas_artificialhomeostaticsystem} showed that the biological machinery for 
homeostasis could successfully be mimicked in a robot platform, laying the groundwork for future research in 
artificial homeostasis systems. The approach taken by \citet{vargas_artificialhomeostaticsystem} was to use a 
combination of artificial neural networks, endocrine networks, and immune systems to adapt the behaviour of a robot 
according to its internal environment. \citet{stradner_analysisimplementationartificial, 
schmickl_modellinghormoneinspired}, on the other hand, use a combination of evolutionary algorithms and diffusion 
systems to mimic the hormonal changes within the human endocrine network, with similar success. While these 
approaches are effective, they often result in cumbersome architectures that are difficult to scale. The {\em 
complicated} systems used by~\citet{neal_timidityusefulemotional} and~\citet{vargas_artificialhomeostaticsystem} are 
limited in how they can approach the {\em complex} problem of autonomous behaviour in robots.

While some changes induced by homeostasis are physiological (sweating, shivering etc.), some are behavioural 
changes. By altering our perception of the world around us, our behaviour is changed to help keep our internal 
environment within certain bounds~\citep{widmaier_vandershumanphysiology}, encouraging certain conscious actions and 
discouraging others. This is the \emph{cognitive} view of homeostasis: the view that cognition is a highly-developed 
form of homeostasis~\citep{lewontin_adaptationspopulationsvarying, ashby_designforabrain, 
godfrey_complexityfunctionmind}. In this paper, rather than viewing homeostasis as the result of interactions 
between the nervous, endocrine, and immune systems, we ask whether a system for artificial homeostasis can be built 
by taking inspiration from cognition.

This work represents a first step towards realising this view of altered perceptions. The robot controller and 
approach we present in this paper provide the ground work for more advanced cognitive experiments. Our approach is 
novel in that we consider the cognitive perspective of homeostasis, and demonstrate that this approach gives rise to a 
much simpler control architecture for homeostatic adaptive behaviour.

In section~\ref{sec:artf_cogn} we describe how we have adapted \cites{cohen_tendingadamsgarden} definition of 
distributed cognition into a robot controller, section~\ref{sec:results} presents a series of experiments 
demonstrating the ability of the robot controller to balance conflicting needs, and so last longer in an environment 
without failing, in section~\ref{sec:discussion} we discuss the impact of the work, and look forward to future 
research in this area, in particular considering how the architecture might be used to alter the perception of the 
world for various forms of robot. The appendix details the methods and experimental setup used for this work, 
including many specific details that are important but distracted from the overall message of the paper.

\section{Implementing Cognitive Homeostasis}
\label{sec:artf_cogn}

While discussing the cognitive nature of the immune system, \citet{cohen_tendingadamsgarden} proposes three 
properties of non-conscious cognitive systems:
 \begin{quote}
   ``Cognitive systems, I propose, differ strategically from other systems in the way they combine three properties:
  \begin{enumerate}
    \item They can exercise options; {\em decisions}.
    \item They contain within them images of their environments; {\em internal images}.
    \item They use experience to build and update their internal structures and images; {\em self-organization}.'' 
(p.~64, emphasis original)

%\citep[p.\,64, emphasis original]{cohen_tendingadamsgarden}

  \end{enumerate}
 \end{quote}

% why this definition? 
% how does this link with homeostasis?

\begin{figure}[h!]
 \centering
 \includegraphics[width=0.6\linewidth]{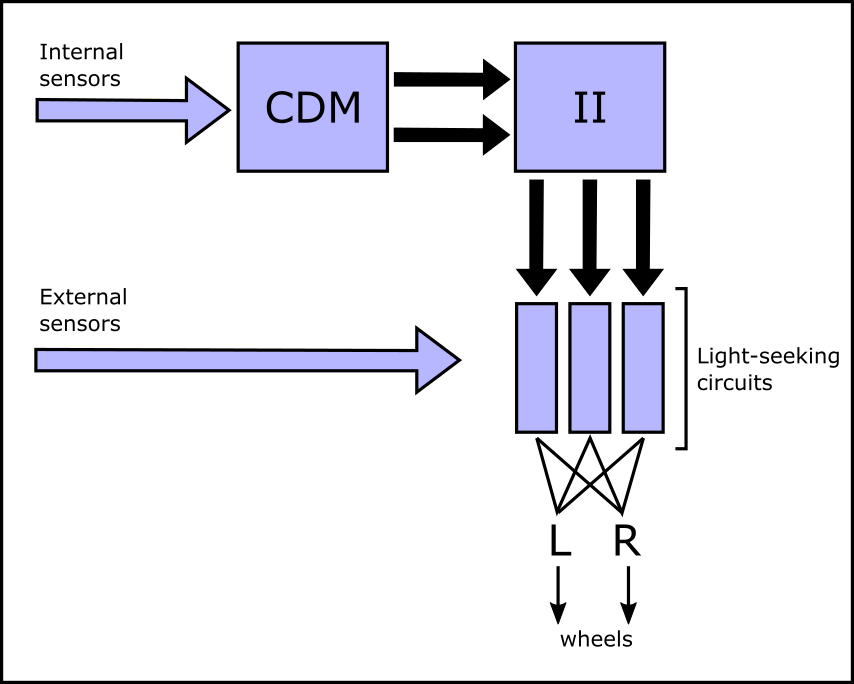}

 \caption[Diagram of the \cah{} system]{High-level architecture of the \cah{} system. The internal sensors (battery 
charge, temperature) drive the behaviour of the \cdmLongLower{} (\cdm{}) module. As the virtual agents within the 
\cdm{} react to the internal sensors, measuring the proportion of agents in each attractor source within the \cdm{} 
can be used to drive the binary signals (black arrow) to the \ii{}'s (\iiLong{}) input. The \ii{} is implemented 
using a form of associative memory known as a Correlation Matrix Memory (CMM). This CMM is trained to recall which 
colour lamp is mounted above charging stations and high-temperature areas) then recalls which colour to search for 
according to the input that is triggered by the \cdm{}. The output from the CMM is then fed into the light-seeking 
circuits that permit the robot to perform basic chromotaxis to seek out or flee from the appropriate colour in the 
environment. }
 \label{fig:cah}
\end{figure}

% details of the overall system and our approach to implementing Cohen's definition
 The \cah{} (\cahLongUpper{}) system, shown in fig.~\ref{fig:cah}, takes Cohen's definition of immunological 
cognition and uses it as the basis for homeostasis in a robot. A robot running \cah{} is able to perform basic 
cognitive tasks: making decisions, learning about the environment, and applying previous experience to influence 
future decisions. These cognitive tasks enable the robot to exhibit homeostatic behaviour and adapt to new 
environments rapidly, an ability not demonstrated by previous approaches~\citep{vargas_artificialhomeostaticsystem, 
stradner_analysisimplementationartificial}.

\cah{} consists of three main components: the \cdm{} (\cdmLong{}) component, the \ii{} (\iiLong{}) component 
(implemented through a Correlation Matrix Memory), and some basic light-seeking circuitry. The \cdm{} runs a novel 
decision-making algorithm (detailed in section~\ref{sec:artf_cogn:dm}), and provides action selection based on the 
robot's internal sensor values.

Over the course of this section, we provide details of how each of Cohen's components for cognition have been 
implemented. Each sub-section is split in two, with the first containing higher-level rationale and description of the 
approach, and the second sub-section providing the low-level implementation details required to fully understand the 
implementation and reproduce the work. Section~\ref{sec:artf_cogn:dm} covers the decision-making component of \cah{}, 
section~\ref{sec:artf_cogn:ii} provides details of the internal memory of \cah{}, and section~\ref{sec:artf_cogn:so} 
presents preliminary results showing how \cah{} learns about its environment.

\subsection{Decision Making}
\label{sec:artf_cogn:dm}

Cognitive systems are typically defined with decision making at the forefront~\citep{cohen_tendingadamsgarden, 
couzin_collectivecognitionanimal, visscher_collectivedecisionscognition, trianni_swarmcognition_interdisciplinary}. 
Decision making is one of the key features of any intelligent system~\citep{pfeifer_understandingintelligence}, which 
may explain its prevalence in definitions of cognition. The ability to change behaviour based on previous experience 
and the current situation allows for the wide repertoire of behaviours that typify cognitive 
systems~\citep{pfeifer_understandingintelligence}. In this paper, we are predominantly interested in decentralised 
systems, especially those where the principles that are applied at the individual level can also scale to the group 
level. Consequently, the most appropriate form of decision making for us to use is collective decision 
making~\citep{bose_collectivedecisionmaking, mchugh_collectivedecisionmaking, marshall_individualconfidenceweighting}.

In order to prevent any prior interpretation of decision-making systems from clouding the behaviour of 
the decision-making algorithm implemented here, we define a collective decision-making system in a 
swarm of agents based directly on~\cites{cohen_tendingadamsgarden} definition:
 \begin{quote}
   ``[A] decision emerges\ldots from a match between an environmental case and an
internal motive. Decisions are associations. \ldots Decision making is positive action; instead of 
passively receiving what the environment imposes, the cognitive system exerts its will\ldots in 
choosing among alternatives'' (p.\,69)
  \end{quote}
 This definition implies that for any given agent (\eg a robot) to make a cognitive decision, it must be able to do more 
than simply react to the environment, it must be able to react differently to the same environmental cues based on 
its internal state.

We define the following terms: 
 \begin{itemize}
   \item an {\em agent} is the entity making a decision, analogous to an `intelligent agent'~\citep{russell_artificialintelligencemodern}. 
 This could be an individual (\eg a robot) or a collection (\eg a swarm of virtual boids).
   \item the {\em context} is the state of the environment as perceived by an agent (Cohen's `environmental case').
   \item the {\em motive} is the likelihood of an agent picking a certain action should an appropriate context arise.
 \end{itemize}

% From these definitions, 
We can see that, in a cognitive system, an agent makes a decision based on some function of context and motive:
 \begin{equation*}
  Decision = f(motive, context)
\end{equation*}

% ignore the robot for a minute -- how do we get a simple agent in an empty environment to make a 
% decision between two options?

To illustrate how motive and context combine to produce decisions, we consider a simple case: a virtual agent in an 
empty simulated environment. The agent can move and can sense the environment around it; we argue here how context 
and motive could be implemented in such an artificial agent.

Let us first consider `context': this is the current state of the environment as perceived by an agent. In the 
simplest case the agent is alone in an empty environment, meaning it should perceive nothing in the environment. If 
we introduce an artefact into the environment (for example, a simple attractor), then the agent will be able to 
perceive changes in its sensors depending on how close to the attractor it is. In our (otherwise empty) simulated 
environment, the attractor can represent one possible decision. We can interpret the agent moving towards the
attractor as indicating a preference of the agent for the particular decision represented by the attractor.

%  \item {\em motive} as the likelihood of picking certain actions should an appropriate context arise. 

Motive is the likelihood of picking an action based on the context. If the agent has the choice of two decisions, 
then it will pick whichever is closest unless the motive pushes it towards some other part of the environment. This 
motive could be as simple as momentum within the agent, but as we are interested in collective decision-making we 
opted for a flock of agents such that a form of `peer pressure' would influence the decision making. This is 
achieved through the `motive' force (see section \ref{sec:artf_cogn:dm:impl} below for details).

Continuing with our example of the simple agent in an empty environment with attractors that represent possible 
decisions, the movement of the agent can be described by a combination of vector adjustments according to input from 
sensors (the `context') and an internal driving force (the `motive'). By using vector adjustments, we are taking a 
similar approach to controlling the agent as that of \citet{reynolds_flocksherdsschools}.

The full details of how we calculate the vector adjustments, along with parameter values and testing are available 
in section~\ref{sec:artf_cogn:dm:impl}. If you wish to skip the finer details then the next part of our artificial 
cognition is described in section~\ref{sec:artf_cogn:ii}.

\subsubsection{Implementation Details}
 \label{sec:artf_cogn:dm:impl}

\begin{comment}

 maths of the flocking system
 details of attractor space and how it is linked with the sensors on the robot	
 how did we test this system in isolation?

 our system has three forces at work:
 - flock force
 - motive force
 - context force
but our description above only covers motive and context, and the flock force should really be covered 
by motive according to our description...

for a flock of agents to make a decision, we need to 
also define motive \todo{I don't like this}at the group level. 

\end{comment}

This section provides details of how the decision-making system is implemented, based on the high-level description 
provided above in section~\ref{sec:artf_cogn:dm}. The decision-making component of \cah{} is used to decide which of 
the low-level signals should affect the high-level behaviour of the robot.

The decision-making component of \cah{} uses a group of virtual agents collected into a flock in a virtual 
environment. The environment is predominantly empty, except for two attractors which provide the context for the 
flock of agents. The environment (and the flock within) will be simulated by the robot, and the actions of the flock 
within its virtual environment will provide decision making to the robot, based on the robot's internal sensor 
values. each attractor in the environment is linked to a sensor on the robot, and the strength of its attraction 
(coefficient of attraction) is varied according to the value arriving from its corresponding sensor.

We have already seen in section \ref{sec:artf_cogn:dm} that a decision results from some function of motive and 
context, given the high-level definitions of what we would consider `context' and `motive'. If the flock is placed 
into an empty environment, where there is no context, the actions of this flock can only result from the inherent 
predisposition of certain actions within the flock, and so---given the lack of context---must be representative of the 
motive of the flock.

As the values of the internal sensors in the robot vary, so do the associated attractors in the virtual environment. 
The flock of agents respond to changes in the virtual environment, linking the behaviour of this virtual flock to 
changes in the internal environment of the robot. The movement of the flock of agents between attractors, therefore, 
provides the action selection required for the high-level behaviour of the robot by reacting to the low-level sensor 
values. In this paper, the attractors are connected to the charge level and temperature sensors on the robot.

\begin{figure}[h!]
 \centering
 \includegraphics[width=0.3\linewidth]{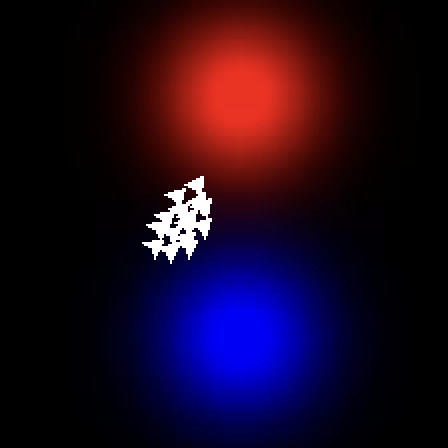}

 \caption[]{Screenshot of simulated decision-making environment embedded into the \cdm{} module. Two attractors are 
 shown in blue and red, representing the decision to charge and to avoid high temperatures, respectively. The flock of 
 agents is moving towards the `avoid high temperatures' attractor, and when more than half the agents are within 7 
 units of the centre, the \cdm{} will notify the \ii{} module that the system should flee high temperatures.}

 \label{fig:artf_cogn:dm:attractors}
\end{figure}

The layout of this virtual environment could be varied in a number of different ways, but---other than varying the 
strength of the attraction---we keep the environment static (dynamic environments is the subject of our ongoing 
research in this field). For this paper, we use an environment of 60x60 units with a periodic boundary condition. The 
attractors are centred at (30, 15) and (30, 45), with a spread of 7 units (see 
fig.~\ref{fig:artf_cogn:dm:attractors}). The flock of agents is initialised at (15, 30). Through the use of emergent 
identities~\citep{stovold_preservingswarmidentity}, multiple flocks could be supported in the same environment but 
this is outside the scope of this paper.

There are many different flocking algorithms that could be used to control our group of agents, most famous of which 
is Reynolds' boids~\citep{reynolds_flocksherdsschools}. For this work, we picked a flocking algorithm that has a more 
rigid internal structure~\citep{olfatisaber_flockingmultiagent}, which allows the information from a contextual cue to 
more quickly propagate through the flock. this also helped to keep the flock cohesive over time, whereas Reynolds' 
boids were more inclined to act as individuals, with the flock breaking up once they encounter an attractor.

The agents in the flock move according to three forces: the \emph{flock} force, the \emph{motive} force, and the 
\emph{context} force. These three forces---described in detail below---combine to provide a final directional force 
that the agents follow. The flock force is provided by Olfati-Saber's 
algorithm~\citep{olfatisaber_flockingmultiagent}, and works to keep the flock in a lattice formation. The motive force 
provides the `momentum' within each agent, and works by projecting `virtual goals' ahead of each agent in the flock. 
The context force provides environmental cues to the agents, resulting from the attractors in the environment.

Combining each of these three forces gives the overall algorithm, described by a series of vector 
adjustments, $u_i$, for an agent $i$:
 \begin{equation}
  u_i = f_i^g + f_i^\gamma
 \end{equation} where $f_i^g$ acts to form a lattice structure among the neighbours in a flock (flock 
force), and $f_i^\gamma$ provides `navigational feedback' towards a goal (combining motive and context 
forces). This definition of $u_i$ varies from \cites{olfatisaber_flockingmultiagent} original 
definition through the removal of the directional term $f_i^d$, relying instead on the navigational 
feedback provided through $f_i^\gamma$.

\paragraph{Flock force} The flock force, $f_i^g$, provides the cohesive force that pushes the agents into a lattice 
formation~\citep{olfatisaber_flockingmultiagent}. It is defined as:
 \begin{equation}
        f_i^g = -\nabla_{q_i}V(q)
 \end{equation} where $V(q)$ is an attractive/repulsive force based on the distance to the nearest neighbouring 
agents, and $q_i$ is the two-dimensional position of agent $i$. $f_i^g$ provides the force required to keep an agent 
$i$ a set distance from its neighbouring agents. The full details of the flocking algorithm are provided in 
\citep{olfatisaber_flockingmultiagent}.

By restricting the agents to only local interactions, the algorithm is applicable to both simulated and real (robotic) 
platforms. In order to provide $f_i^\gamma$ to the flock without global information, each agent projects a `virtual 
goal' ahead of itself, which collectively provides the navigational feedback required to prevent the flock from 
disintegrating \citep{olfatisaber_flockingmultiagent}. This virtual goal is the key component of the `motive force'.

\paragraph{Motive force} The motive force is provided through a `virtual goal' in the environment. This is calculated 
by each agent projecting forward by a predetermined distance, \param{d}, and using that position in the environment as 
its goal for a set period of time, \param{G$^{\param{update}}$}. These parameters, \param{d} and 
\param{G$^{\param{update}}$}, are set at the start of each simulation run, and do not vary throughout the run or 
between agents. Between them, these parameters provide a weighting towards motive or context for the agents (\ie a 
weighting between the motive force and the context force).

Specifically, an agent $i$ calculates its virtual goal, $G_i$, by taking the
average heading of its neighbours, $N_i$, within a pre-defined radius, \param{r}.
Let the average heading of the neighbours of agent $i$ be $\phi_i$, defined as:
 \begin{equation}
   \phi_i = \dfrac{1}{|N_i|} \sum_{j\in N_i} \theta_j
 \end{equation}

 \noindent then projecting forwards by the predefined distance, \param{d}, gives the virtual goal for an agent $i$ as 
 the coordinate pair $G_i = (x_{G_{i}},y_{G_{j}})$, providing the motive force $f_i^{motive}$:
\begin{align}
   x_{G_{i}} =&\, x_i + d\cdot cos(\phi_i) \\
   y_{G_{i}} =&\, y_i + d\cdot sin(\phi_i)
 \end{align}
 The goal is recalculated periodically so that the motive reflects the current state of the agent, including 
influences from the environment. This update period is parameterised as the virtual goal update interval 
(\param{G$^{\param{update}}$}). As the interval is increased, the agent is weighted further towards the motive than 
the context (and vice-versa), as information from the context will influence the position of the virtual goal less 
often.

\paragraph{Context force} The environment is empty other than any attractor sources included. An attractor, $j$, 
exerts a force on all agents in the environment. The force at any point can be described by the Gaussian probability 
density function, with mean $\mu_j$, standard deviation $\sigma_j$, and distance to the centre of the attractor from 
agent $i$, $(q_i - \mu_j)$.
 \begin{equation}
 \label{eqn:gaussian}
  f_i^{context}(\mu_j,\sigma_j,q_i) = \dfrac{1}{\sqrt{2\sigma_j^2\pi}}\, e^{-\dfrac{(q_i - \mu_j)^2}{2\sigma_j^2}}
 \end{equation}
 The agent calculates $f_i^\gamma$ based on the coordinates of $G_i$ and the
gradient formed by the attractors in the environment:
 \begin{equation}
 \label{eqn:fgamma}
  f_i^\gamma = f_i^{motive}(q_i,p_i,q_{G_i},p_{G_i}) + f_i^{context}(\mu_j,\sigma_j,q_i)
 \end{equation}
 \noindent Each agent multiplies the attractor gradient by a global parameter
\param{ctx$\_$mult} as it senses it, in order to provide a weighting between
$f_i^{motive}$ and $f_i^{context}$. See table \ref{tab:dm:gamma} for a summary of
each term used in the definition of $f_i^\gamma$.

\begin{table}[h!]
 \centering
 \begin{tabular}{l|l}
  \textbf{Variable}             & \textbf{Description}                          \\ \hline\hline
  $q_i$                         & Two-dimensional position of agent $i$         \\ \hline
  $p_i$                         & Velocity of agent $i$                         \\ \hline
  $G_i$                         & Position of virtual goal for agent $i$        \\ \hline
  $\mu_j$                       & Centre-point of attractor $j$                 \\ \hline
  $\sigma_j$                    & Spread of attractor $j$                       \\
 \end{tabular}
 \caption[Variables used in definition of $f_i^\gamma$]{Descriptions of the variables used in equation 
\ref{eqn:fgamma}.}
 \label{tab:dm:gamma}
 \end{table}

\begin{table}[tbh]
 \centering
 \begin{tabular}{l|l|l|l}
  \textbf{Parameter}                    & \textbf{Description}                          & \textbf{Value}  & \textbf{Units}  	\\ \hline\hline
  \param{ctx$\_$mult}                   & Multiplier for attractor gradient             & 185   & N/A		\\ \hline
  \param{d}                             & Distance from agent to virtual goal           & 30    & patches	\\ \hline
  \param{G$^{\param{update}}$}  	& Update period for virtual goal                & 25    & timesteps	\\ \hline
  \param{r}                             & Radius used to calculate neighbourhood        & 8     & patches 	\\
 \end{tabular}
 \caption[Typical parameter values for the decision-making simulation]{Typical parameter values and 
descriptions for the decision-making simulation. `Patches' and `timesteps' are generic terms for 
coordinate space in the simulation and discretised simulated time, respectively. }
 \label{tab:parameters}
\end{table}

%\notejhs{ do we include any testing or demonstration that the DM system actually works? never did in 
%the frontiers paper, but then again that paper was rejected, so\ldots }

\subsection{Internal Images / Memory}
\label{sec:artf_cogn:ii}

%% need more motivation for this section. not just `cohen says so', but why is it important? why do we 
%% need it for a cognitive system?

%    They contain within them images of their environments; {\em internal images}.
%   ``Internal images carve reality into bite-size [pieces]'' (p.\,70)

\citet{cohen_tendingadamsgarden} introduces memory through the concept of an `internal image'. This is a functional or 
physical imprint of the external environment on the cognitive system. These internal images provide a way for an 
otherwise blind system to interpret and interact with the outside world, translating from the internal `language' of 
the cognitive system to actions in the external world. It is from this perspective that we have implemented memory in 
our system, to translate the decisions from the \cdm{} module into actions in the real world. If the \cdm{} module 
says `we need to charge', the \iiLong{} (\ii{}) module recalls the last environmental cue in which the robot was 
charged. For example, if this was a blue light then the \ii{} module will output `seek out blue light'.

% \citeyear{palm_towardstheorycellassemblies})
% apa bibstyle is messing up the dual citation in this paragraph, hence the mess in the citep command:

The \ii{} module lends itself well to associative memory: a mechanism for associating stimuli with responses. In 
biological systems, the memory results from temporal associations that emerge between two sets of interacting 
neurons~\citep[\citeyear{palm_towardstheorycellassemblies}]{palm_onassociativememory}. As the stimulus is presented to 
the first set of neurons, the response of those neurons is sent to the second set of neurons, which subsequently 
respond. This second response is similar every time, indicating that the same response will occur if given the same 
stimulus.

CMMs (Correlation Matrix Memories)~\citep{kohonen_correlationmatrixmemories} are one type of associative memory that 
have a particularly simple structure, a fast response time, and they can be trained in a single pass, meaning 
associations can be built up as the robot experiences new parts of the environment. These properties make them ideal 
for our purposes in the \ii{} module.

Section~\ref{sec:artf_cogn:ii:impl} provides an overview of CMMs along with details of how we use CMMs in the \ii{} 
module to translate between the internal language of the cognitive system and actions in the real world.

\subsubsection{Implementation Details}
\label{sec:artf_cogn:ii:impl}

In this section we introduce an overview of Correlation Matrix Memories (CMMs) and detail how they are 
used in the \cah{} system through the \iiLongLower{} module.

\begin{figure}[h!]
 \centering

 \subfloat[Basic CMM architecture (left), where $\mathcal{M}$ represents the matrix of binary weights, $\mathcal{I}$ 
and $\mathcal{O}$ represent the input--output vector pair corresponding to the neurons $a,b,c$ and $d,e,f$ in (b), 
respectively. \label{fig:cmm:maths}]{ %
 \centering
 \parbox{0.5\linewidth}{ \vspace{-12em} %

 \begin{displaymath}
 \begin{array}{lc|cr}
 \begin{matrix}
   \begin{pmatrix}
     ~ \\
         \mathcal{I} \\
         ~
    \end{pmatrix}
   &
        \begin{pmatrix}
         ~ & ~ & ~ \\
         ~ & \mathcal{M} & ~ \\
         ~ & ~ & ~
        \end{pmatrix}
   \\
        ~ \\
   &
        \begin{pmatrix}
         ~ & ~\mathcal{O} & ~~
        \end{pmatrix}
  \end{matrix}
 & ~~ & ~~ &
  \begin{matrix}
    \begin{pmatrix}
        a \\
        b \\
        c
        \end{pmatrix}
  &
        \begin{pmatrix}
      ad & ae & af \\
          bd & be & bf \\
          cd & ce & cf
        \end{pmatrix}
  \\
        ~ \\
  &
        \begin{pmatrix}
          ~d & ~e & ~f~
        \end{pmatrix}
  \end{matrix}
 \end{array}
 \end{displaymath}  }
} 
 \subfloat[CMM associative memory neural network. The binary weights between the
two layers are represented by the matrix $\mathcal{M}$ in (a). \label{fig:cmm:nn}]
{ \centering %
  \includegraphics[width=0.4\linewidth]{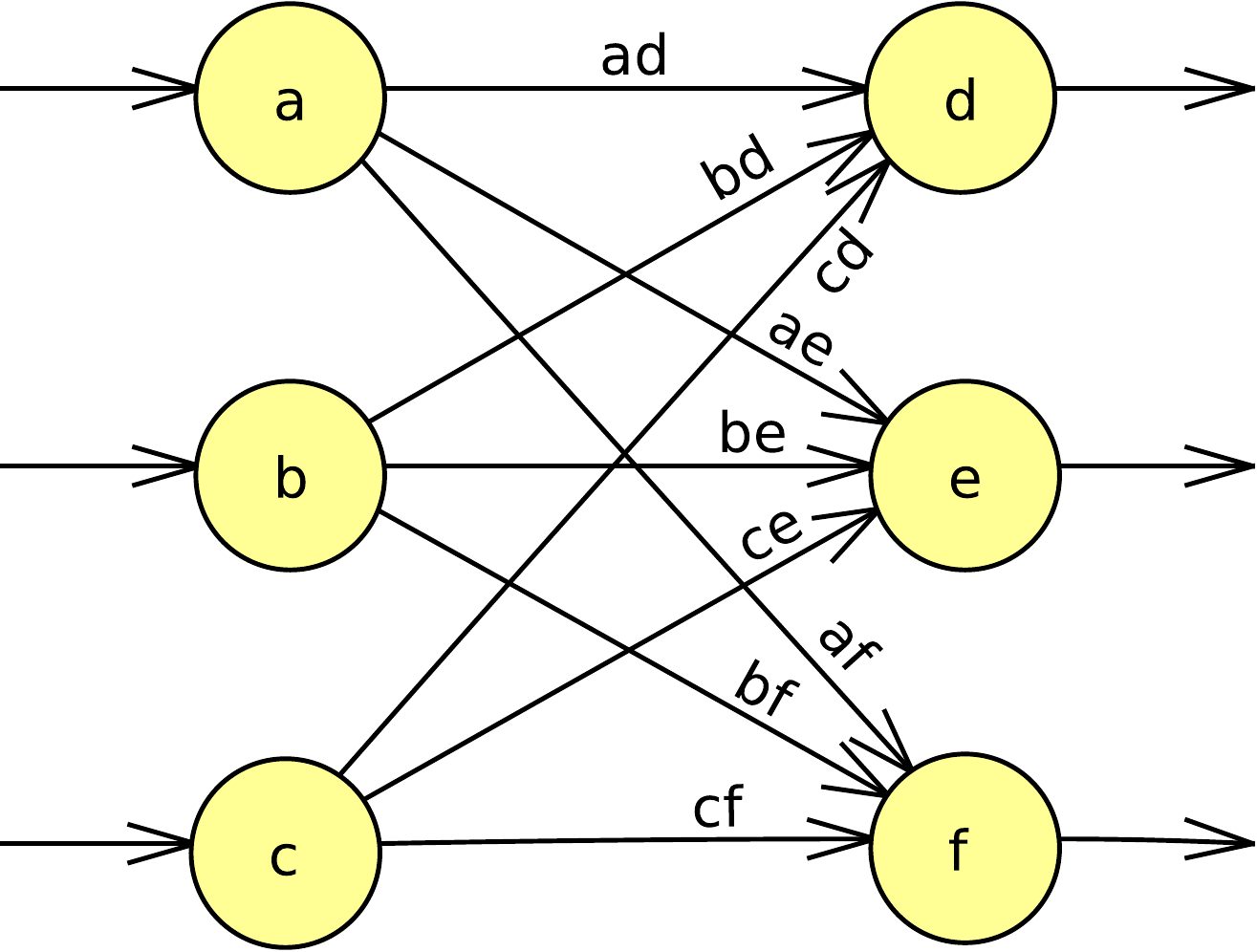} }
 \caption[CMM associative memory neural network]{}
 \label{fig:cmm}
\end{figure}

The CMM (Correlation Matrix Memory)~\citep{kohonen_correlationmatrixmemories} is a matrix-based representation of a 
two-layer binary neural network. The matrix represents the binary weights from a fully-connected, two-layer artificial 
neural network (one input layer, one output layer; see fig.~\ref{fig:cmm:nn}). For example, the network in 
fig.~\ref{fig:cmm:nn} would be represented by the CMM, $\mathcal{M}$, with $k$ input--output pairs $\mathcal{I}$ and 
$\mathcal{O}$ in fig.~\ref{fig:cmm:maths}.

Before training, the initial matrix $\mathcal{M}$ is filled with zeros (as there are no associations 
stored in the network). As the $k$ binary-valued input--output pairs are presented to the network, 
associations are built up in the matrix $\mathcal{M}$. These associations are stored as 1s in the 
matrix, corresponding to coincident 1s in both input and output vectors. For example, in 
fig.~\ref{fig:cmm:maths}, if $a$ and $e$ were both 1 then $ae$ would be set to 1 after training.

In the \ii{} module, we use a CMM to associate changes in the environment with
changes in internal motive, following on from \cites{cohen_tendingadamsgarden}
definition of cognitive decision-making:
 \begin{quote}
   ``[A] decision emerges\ldots from a match between an environmental case and an
internal motive. Decisions are associations.'' (p.\,69)
 \end{quote}
 \noindent He continues:
 \begin{quote}
   ``Decision-making is positive action; instead of passively receiving what the
environment imposes, the cognitive system exerts its will\ldots in choosing among
alternatives''~(p.\,69)
  \end{quote} 
The associations made between the environmental case (referred to throughout this paper as `context') and internal 
motive are made explicitly in the \ii{} module, but also implicitly in the \cdm{} module described in 
section~\ref{sec:artf_cogn:dm}.

The output from the \cdm{} module (\ie the robot motive) is passed as an input to the CMM. In our system, the \ii{} 
module makes use of a CMM to store its `internal image'---or imprint---of the environment for the robot to use. The 
\ii{}, therefore, provides the robot with the capacity to make associations between changes in internal and external 
sensor values that occur at the same time. For example, if we mark a charging station with a blue light, the robot can 
sense the increase in blue light at the same time as an increase in battery charge. By associating these two signals, 
the \ii{} offers the ability to recall which colour to search for in order to find a charging station.

Fig.~\ref{fig:cmm:sensors} shows how the \ii{}'s CMM is constructed for two internal sensors (charge, $c$, and 
temperature, $t$) and for the three colour components from an external light sensor (red, green, and blue). When one 
of the sensors passes a threshold, the corresponding binary value switches from 0 to 1. If this occurs on one of the 
internal sensors at the same time as one of the external sensors, then the corresponding value in the association 
matrix is set to 1 as well, storing this association between the two sensors. Re-presenting one of the internal sensor 
inputs (\eg `charge'), by setting it to 1 again will recall the association. This will cause the closest stored 
associations to be recalled, and the appropriate values set in the output vector. This allows the system to test which 
colours have been associated with which inputs.

\begin{figure}[h!]
 \centering
\begin{displaymath}
\begin{array}{lc|cr}
 \begin{matrix}
  \begin{pmatrix}
    c \\
    t
  \end{pmatrix}
 &
  \begin{pmatrix}
        cr & cg & cb \\
        tr & tg & tb
  \end{pmatrix}
 \\
  ~ \\
 &
  \begin{pmatrix}
    ~r & ~g & ~b~
  \end{pmatrix}
 \end{matrix}
 & ~~ & ~~ &
 \begin{matrix}
   \begin{pmatrix}
         1 \\
         0
   \end{pmatrix}
&
   \begin{pmatrix}
        ~0 & ~0 & ~1 \\
        ~0 & ~0 & ~0
   \end{pmatrix}
  \\
        ~ \\
  &
        \begin{pmatrix}
        ~0 & ~0 & ~1~
        \end{pmatrix}
 \end{matrix}
\end{array}
\end{displaymath}
 \caption[Basic CMM for the \cah{} system]{Basic CMM architecture (left), where $c$ and $t$ correspond to the internal 
sensors for charge and temperature, and $r$, $g$, and $b$ correspond to the red, green, and blue colour components 
from the light sensor, respectively. The vectors $\icol{c\\t}$ and $(r\,g\,b)$ are provided as the stimulus--response 
pair of vectors to the association matrix. Any points in the association matrix that has a 1 on both input vectors is 
set to 1, storing the association between the two. An example association is shown on the right, as would occur after 
the robot finds a charging station under a blue light.}
 \label{fig:cmm:sensors}
 \end{figure}

\subsection{Self-organization / Learning}
\label{sec:artf_cogn:so}

\cites{cohen_tendingadamsgarden} final property of cognitive systems is: ``They use experience to build 
and update their internal structures and images; {\em self-organization}.'' (p.\,64). From Cohen's 
viewpoint of immunological cognition, self-organization and learning look very similar, in fact he 
subsequently defines self-organization as: 
\begin{quote}
 ``The hallmark of cognitive self-organization is the process we call {\em learning}; a cognitive 
 system, through experience, acquires new capabilities and behaviors.'' \citep[p.\,82; emphasis 
 original]{cohen_tendingadamsgarden}
\end{quote}

\noindent With this in mind, in this section we describe how the system can learn, and how it is set up to enable 
learning in new ways in future.

There are two main areas with the capacity to learn in the \cah{} system: the \ii{} and the \cdm{}. The \ii{} is the 
only area which we allow to learn for the work presented here, and the learning takes place through the association of 
internal needs and external sensor changes (\ie when the robot is under a blue light and charging). This learning is 
sufficient for the work we are presenting in this paper, but there are other, more complex ways of learning in \cah{}.

The \cdm{} relies on the use of attractors that are linked to the internal sensors on the robot. With a fixed number 
of internal sensors (in this case temperature and charge), it was logical to fix the number of attractors in the 
virtual environment. There is, however, no reason why the \cah{} system must be used in this way: we could link the 
\cdm{} to a visual system which provides a much wider range of stimuli to the \cdm{}. If this were the case, the 
argument for a dynamic virtual environment is much stronger. This is the focus of our current research efforts, but is 
far outside the scope of this paper.

\subsubsection{Implementation Details}
\label{sec:artf_cogn:so:impl}

 The \cah{} system enables the robot to adapt to its environment through the use of the \ii{}'s CMM. The CMM 
associates large changes in internal sensor values with large changes in external sensor values. For example, if the 
robot discovers a charging station under a green lamp, the CMM will associate green light with charging. While, in 
principle, the \cah{} system is able to adapt on-line, this functionality is disabled in all experiments presented in 
this paper. This ensures that any variation between test and control cases are only due to swapping out the \cdm{}. 
This section aims to demonstrate that a robot running the \cah{} system can learn from its experience in an 
environment, and put it into practice in a simulated environment ahead of being used in real robot experiments.

\begin{figure}[h!] 
 \centering 
 \includegraphics[width=0.45\linewidth]{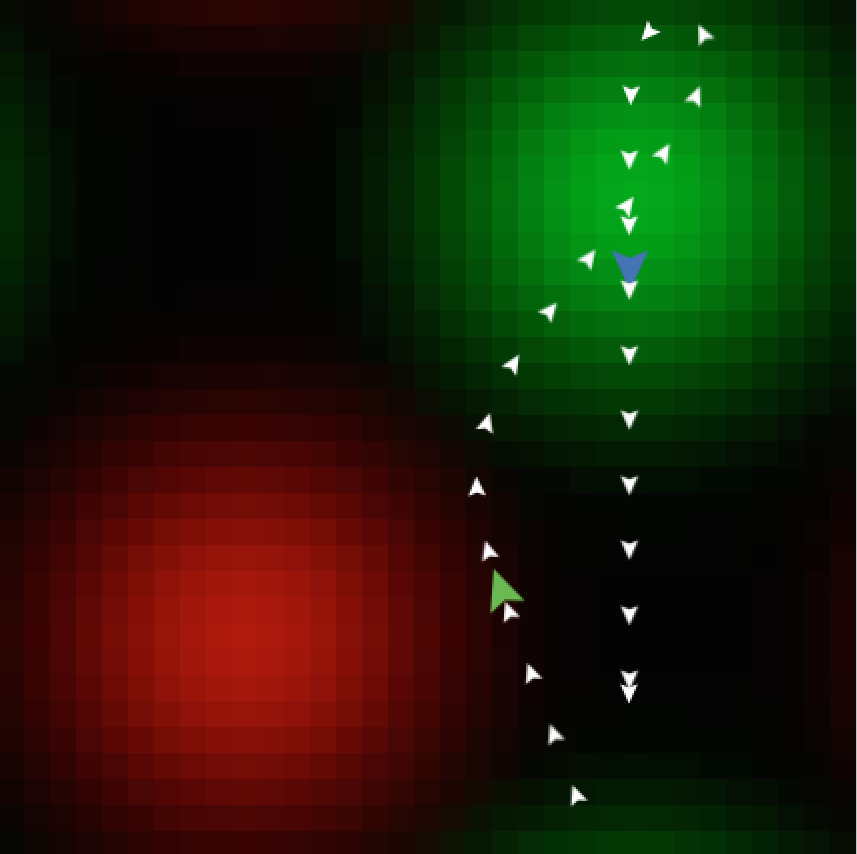}
 \caption[Screenshot of the proof-of-principle simulation]{Screenshot of the proof-of-principle simulation, along with 
 trace of the post-training robot searching for the green light source. The white arrows are the trace of the robot 
 position, the green and blue arrows signify when the `needs charge' motive switch is enabled and disabled.} 
 \label{fig:cah:simulator}
\end{figure}

 The simulation of our robot is shown in fig.~\ref{fig:cah:simulator}. This simulation is constructed in 
NetLogo~\citep{wilensky_netlogosimulator} and has the entire \cah{} system implemented. The signals provided by the 
\cdm{} to the rest of the system, however, are provided using switches on the simulator interface. The simulation 
consists of a single, zero-mass robot (a high-level representation of the real robot running \cah{}) that roams 
around a simple environment with different coloured lights (red and green, in this case). In place of the \cdm{}, we 
provide a mechanism for signalling that the robot has a certain need. This `need' is usually provided by the 
decision-making ability of the \cdm{}. By signalling that the robot has a certain need, the \cah{} system provides a 
mechanism for the robot to find the corresponding part of the environment. For example, `needs charge' recalls the 
appropriate colour of light from the CMM and searches for that colour in order to recharge.\footnote{we don't 
provide the literal term `needs charge' to the \cah{} system, the system instead provides a signal that indicates it 
is running low on charge.}

 \paragraph{Simulated training} The simulated robot roams around the environment during its training phase, learning 
about the environment. Once this has been completed the system is ready to be tested. In this case, if the training 
is successful, the robot should head towards the green light when the `needs charge' motive is switched on.

 Fig.~\ref{fig:cah:simulator} shows a trace of the simulated robot moving around the environment. Once the `needs 
charge' switch is enabled (green arrow), the robot heads towards the green light source. Once this motive is 
disabled (blue arrow), the robot leaves the light source. This shows that, in principle, the system has the capacity 
to learn about its environment.

 % The simulation shows that the architecture functions as it should when appropriate signals are
 % provided from the \cdm{} component (which is not included in this simulation). This implies that any
 % variation we see by including a \cdm{} component (or a control component) can only be from the
 % changes within the \cdm{}, or noise in the environment, rather than as a result of the architecture.

 The post-training behaviour of the simulation shows that the \cah{} system is able to search for a specific part of 
the environment, based on high-level inputs such as a `needs charge' signal. The system translates this signal back to 
a colour, based on previous experience of the environment. In other words, the system recalls `green' as it learned 
that its batteries were charged when under a green light, which it subsequently seeks out.

% At this stage, the `motive' of the agent is represented by a switch that is controlled through the 
% UI, but this `motive' will be provided by the \cdm{} algorithm when implemented into a robot.

 Having shown that the system behaves as expected in a simulated environment, the next step is to implement this 
system on a hardware robot platform, and test whether this behaviour is consistent.

 \paragraph{Robot-based training} The \cah{} system is implemented onto a robot (details of the robot platform and 
experimental setup are provided in appendix~\ref{app:setup}). As before, the robot is allowed to roam around the 
environment during its training phase, learning about the environment. The environment (shown in 
fig.~\ref{fig:robot:cdm:arena}) consists of two coloured lamps (one red, one blue), and the \cah{} is set up to 
simulate charging its batteries under the blue light, and to simulate an increase in temperature under the red light.

% In the training phase, the robot is trained using a random walk around the environment from an
% arbitrary start point.
%  The threshold for external (colour) sensor values is 600 acu; for charging stations, 500 acu; for 
%  high-temperature areas, 300 acu (see appendix~\ref{app:setup} for details of the colour unit `acu'). 

 As the robot roams, the CMM associates high values from the internal sensors with high values from the external 
 sensors. Once the robot has roamed across both areas of the environment with lamps in, the robot is taken from the 
 arena and the contents of the CMM downloaded.

 Due to the small size of the CMM used in this paper (3x2 matrix), the training phase can be tested
with relative ease. In the following matrices, the following key can be used to determine whether the
training has been successful, $r,g,b$ correspond to the red, green, and blue external sensors, and
$c,t$ correspond to the charge and temperature internal sensors:
 \begin{displaymath}
 \begin{matrix}
   \begin{pmatrix}
        cr & cg & cb \\
        tr & tg & tb
   \end{pmatrix}
 \end{matrix}
\end{displaymath}
 
\noindent After exposing the robot to an environment with one blue lamp over a charging station, the CMM is trained 
correctly as:

 \begin{displaymath}
 \begin{matrix}
   \begin{pmatrix}
        ~0 & ~0 & ~1 \\
        ~0 & ~0 & ~0
   \end{pmatrix}
 \end{matrix}
\end{displaymath}

\noindent After exposing the robot to an environment with both a red lamp over a high-temperature area and a blue lamp 
over a charging station, the CMM is trained correctly as:

\begin{displaymath}
 \begin{matrix}
   \begin{pmatrix}
        ~0 & ~0 & ~1 \\
        ~1 & ~0 & ~0
   \end{pmatrix}
 \end{matrix}
\end{displaymath}

% While there is not space in this paper to fully detail the training results, the simplicity of these
% results should give the reader confidence that this architecture is able to learn about its
% environment with ease. These training results are subsequently used as the basis for the experiments
% presented below.

\section{Results}
\label{sec:results}

%\notejhs{this entire section has just been copied across from the Frontiers paper, with the exception 
% of fixing a glaring error in the results table. The RQs need updating according to how we present it}

 \begin{minipage}{\linewidth}
\noindent By comparing with a control case, we demonstrate that \cah{} can\ldots
\begin{quote}
\begin{itemize}
 \item[\textbf{\resQu{1}}:] \ldots enable a robot to learn from previous experiences and use them to influence future behaviour
 \item[\textbf{\resQu{2}}:] \ldots provide homeostatic behaviour to a robot
 \item[\textbf{\resQu{3}}:]  \ldots balance two conflicting requirements to provide homeostatic behaviour to a robot 
\end{itemize}
\end{quote}
 \end{minipage}
 \newline\newline 
 This control case uses a replacement for the \cdm{} that reacts to its internal state directly with a fixed threshold. 
The \ii{} and light-seeking circuits shown in fig.~\ref{fig:cah} are still present and functioning, and the \cdm{} module 
still operates in order to take up the same CPU and memory time but without having any influence on the rest of the 
system.

\subsection{\cah{} provides homeostatic behaviour to a robot}
\label{sec:robot:results:two_sensor}

This section considers how a robot running \cah{} is able to alter its high-level behaviour according to its low-level 
needs. As the internal state of the robot varies, the \cdm{} signals to the rest of the \cah{} architecture what it 
needs in order to keep the state within its limits.

% The internal state for the robot platform is markedly different to that of biological systems, but
% providing even limited information to the \cdm{} is sufficient for it to guide the behaviour of the
% robot in the environment.

The \cdm{} works to keep the internal sensors (charge and temperature) within certain limits. If the temperature gets 
too high or the charge too low then the robot will fail. The sensors are simulated so that the internal behaviour of 
the robot can be measured (if the battery actually ran out of charge, the data about internal behaviour might be lost 
or corrupted). The values used for different variables in our experimental setup are given in 
table~\ref{tab:robot:internal_sensors}.

\begin{table}[h!]
 \centering
 \begin{tabular}{r|c|c}
        ~                               & \textbf{Temperature}  & \textbf{Charge} \\ \hline\hline
        {Start value}             & 10.0                                  & 5.0           \\ \hline
        {+ve Delta}               & 1.8                                   & 0.8           \\ \hline
        {-ve Delta}               & 0.5                                   & 0.1           \\ \hline
        {Fail Point}              & --                                    & 0.1           \\ \hline
        {Limits}                  & [10.0, 60.0]                          & [0.01, 15.0]            \\ \hline
        {Control threshold}       & --                                    & 2.5
 \end{tabular}
 \caption[Parameter values for the basic homeostasis experiment]{Parameter values used in the basic
 homeostasis experiment. The values have been picked to result in a challenging, but possible scenario. }
 \label{tab:robot:internal_sensors}
\end{table}

\begin{figure}[h!]
 \centering
 \includegraphics[width=0.7\linewidth]{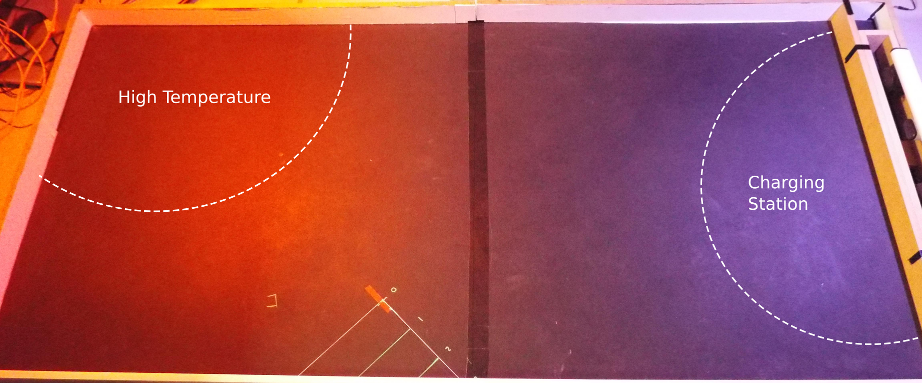}
 \caption[Arena for homeostasis experiment]{Photo of arena with charging and warm areas superimposed, arena size: 224x108cm}
\label{fig:robot:cdm:arena}
\end{figure}

This section looks at a basic scenario: asking whether the robot is able to keep itself charged while attending to 
another task. This other task is to warm itself under a lamp. By arranging the experimental arena (depicted in 
fig.~\ref{fig:robot:cdm:arena}) so that the charging station (a blue lamp) is far away from the warming lamp (a red 
lamp), the robot needs to leave the warming lamp in order to recharge, and vice-versa. This should result in the robot 
switching from sitting under one lamp to sitting under the other repeatedly and indefinitely. The parameters are 
chosen such that it is not able to warm itself sufficiently to complete its task before needing to charge again~(thus 
cooling the robot back down again). This means that the robot will eventually run out of charge, but we are interested 
in how long the robot is able to balance the two tasks.

\nhypothesis A robot running the \cah{} architecture will survive no longer with the \cdm{} component
than with the threshold component.\newline

The design of the system described in section~\ref{sec:artf_cogn} was such that the \cdm{} could be replaced by 
another component that provides different signals, based on the values of the internal sensors. The control case uses 
a simple threshold mechanism to achieve this and provide a viable alternative to the \cdm{}. The control output 
switches from 0 to 1 once the charge value drops below the control threshold of 2.5 (see appendix for calculation of 
appropriate control threshold values).

%simulation results

\begin{figure}[h!] 
\centering 
 \subfloat[Simulation environment, with white arrow representing the robot, blue light and red light representing the 
 charging station and warming area, respectively. \label{fig:robot:results:sim_homeo:setup}] %
     {\centering  \includegraphics[width=0.5\linewidth,valign=c]{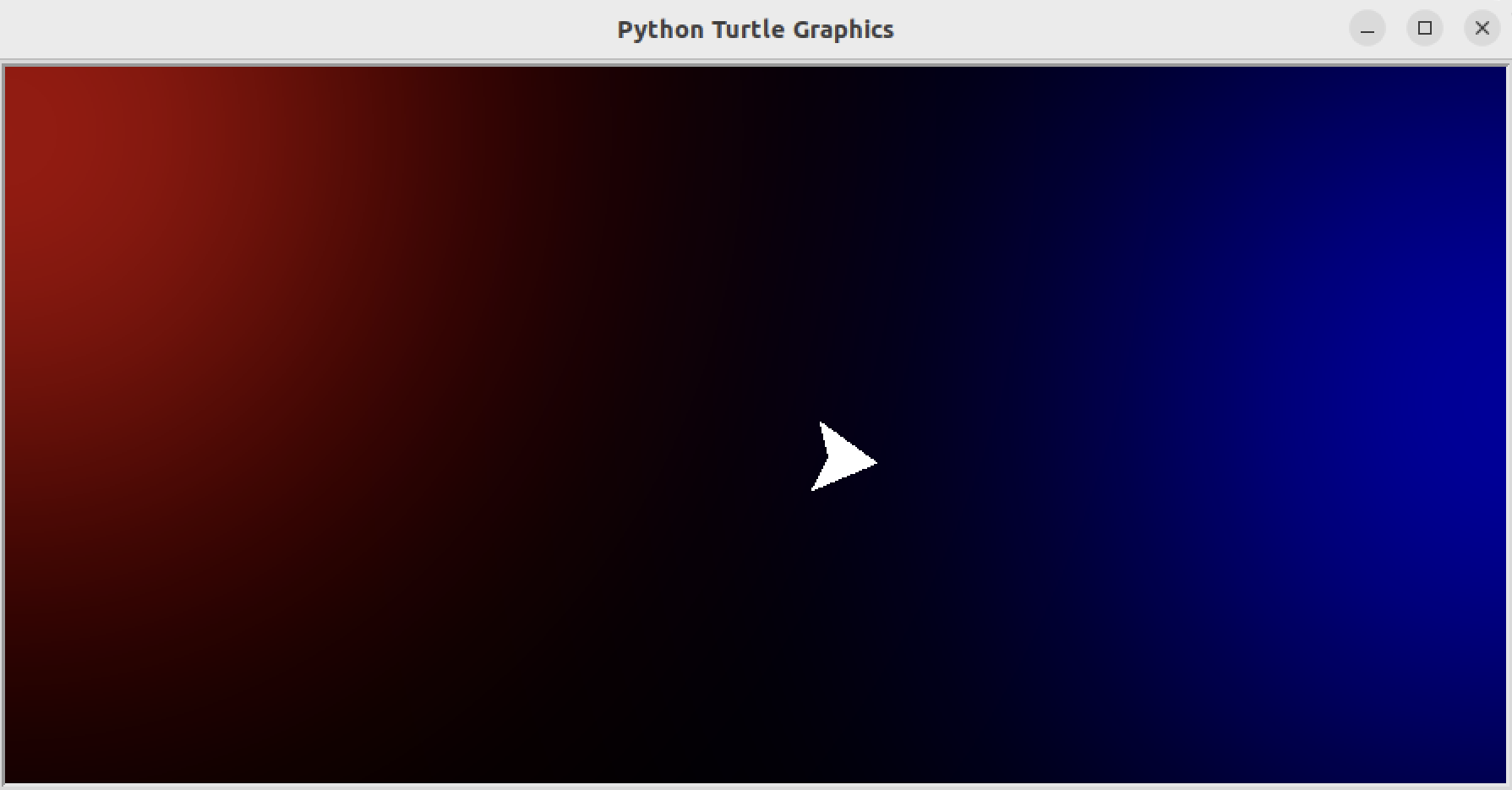} %
       \vphantom{ \includegraphics[width=0.4\linewidth,valign=c]{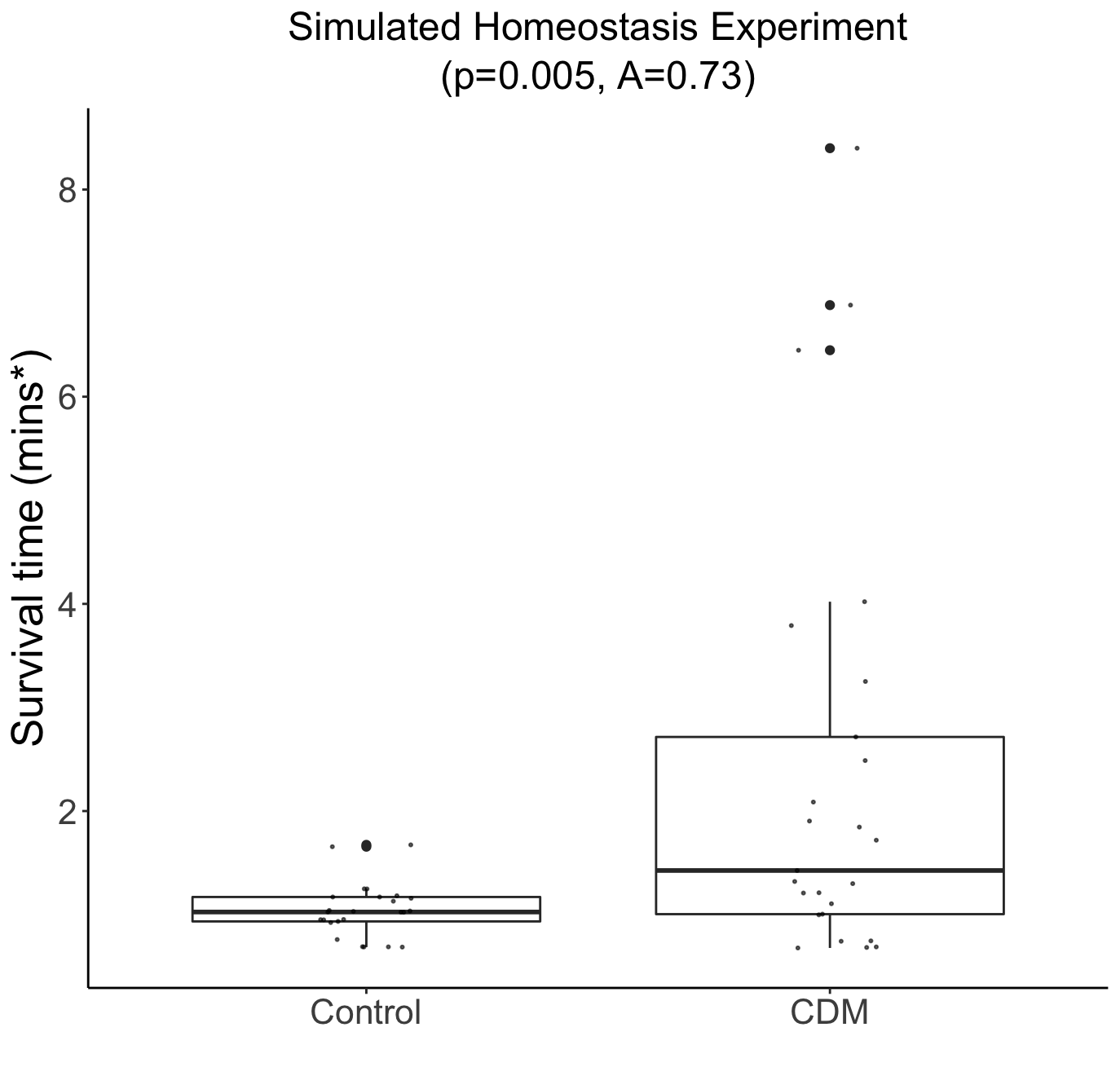} }} \quad
 \subfloat[Boxplots showing that the \cdm{} enables the robot to survive for significantly longer than a simple 
 threshold controller (higher is better). The Mann--Whitney--Wilcoxon test and Vargha--Delaney A-test give $p = 0.005$ 
 and $A = 0.73$ respectively, rejecting $H_0$ in simulation with 95\% confidence.
 \label{fig:robot:results:sim_homeo:boxplots}] %
 { \includegraphics[width=0.4\linewidth,valign=c]{sim_homeostasis.png}}

 \caption[]{Environment setup and results from simulated experiment (mimicking the real-world setup). N.B.\ the 
 survival time is measured in minutes, but the simulation runs faster than the real robot, so the results are only 
 proportionally comparable to the results gathered from real robots (presented below).}

 \label{fig:robot:results:sim_homeo}
\end{figure}

Due to the amount of time that robot lab experiments take, we constructed a simulated environment that mirrors the lab 
setup (see fig.~\ref{fig:robot:results:sim_homeo:setup}). Through this simulation, we are able to demonstrate the 
behaviour of the system in an idealised environment, before confirming the behaviour in a real-world lab setup. The 
simulation uses the same C++ code as the real robot, wrapped in Python and using Turtle graphics to represent the 
environment and robot. This code is available through the Github repository linked at the top of this 
article.

Fig.~\ref{fig:robot:results:sim_homeo:boxplots} shows the time for which a simulated robot is able to survive in the 
environment with the CDM and control setups. The data show a clear distinction between the \cdm{} and control case, 
as supported by the statistical tests showing a significant difference between the distributions. 

\begin{figure}[h!] \centering \subfloat[Trace of robot path between 5.5 and 10.5 mins superimposed on the environment. 
 The blue and red circles indicate points where the robot stopped to charge or warm itself respectively. The green and 
 white circles indicate the start and end points. The red and blue solid lines indicate that the robot is searching 
 for that colour. The white dotted line is when the robot wanders (not actively searching for anything).  
 \label{fig:robot:results:example_graphs:trace}] { \centering % 
 \includegraphics[width=.8\linewidth]{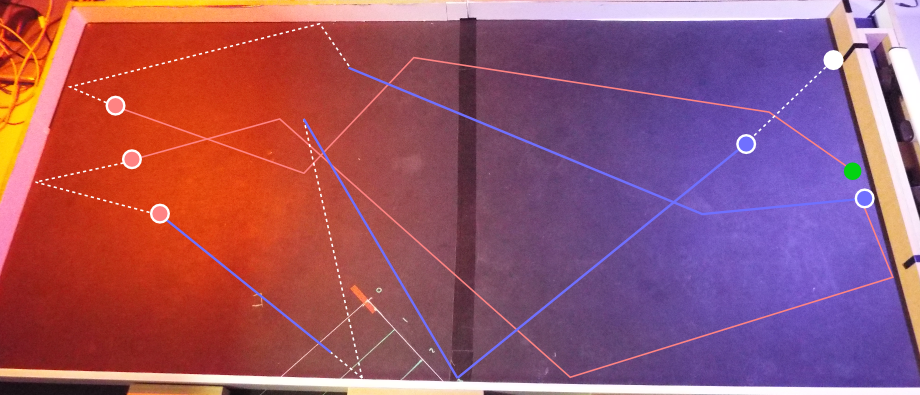} } \\ 
\subfloat[Graphs showing internal sensor values (top)  and \cdm{} outputs (bottom) for the robot between 5.5 and 10.5 mins. 
 The red and blue line show the internal values for temperature and charge as presented to the \cdm{} component. The 
 red and blue blocks of colour in the bottom graph show the output of the \cdm{} component (red indicating to the 
 robot to search for red light, and blue indicating to search for blue light).  
 \label{fig:robot:results:example_graphs:graphs}] { \centering %
 \includegraphics[width=.7\linewidth]{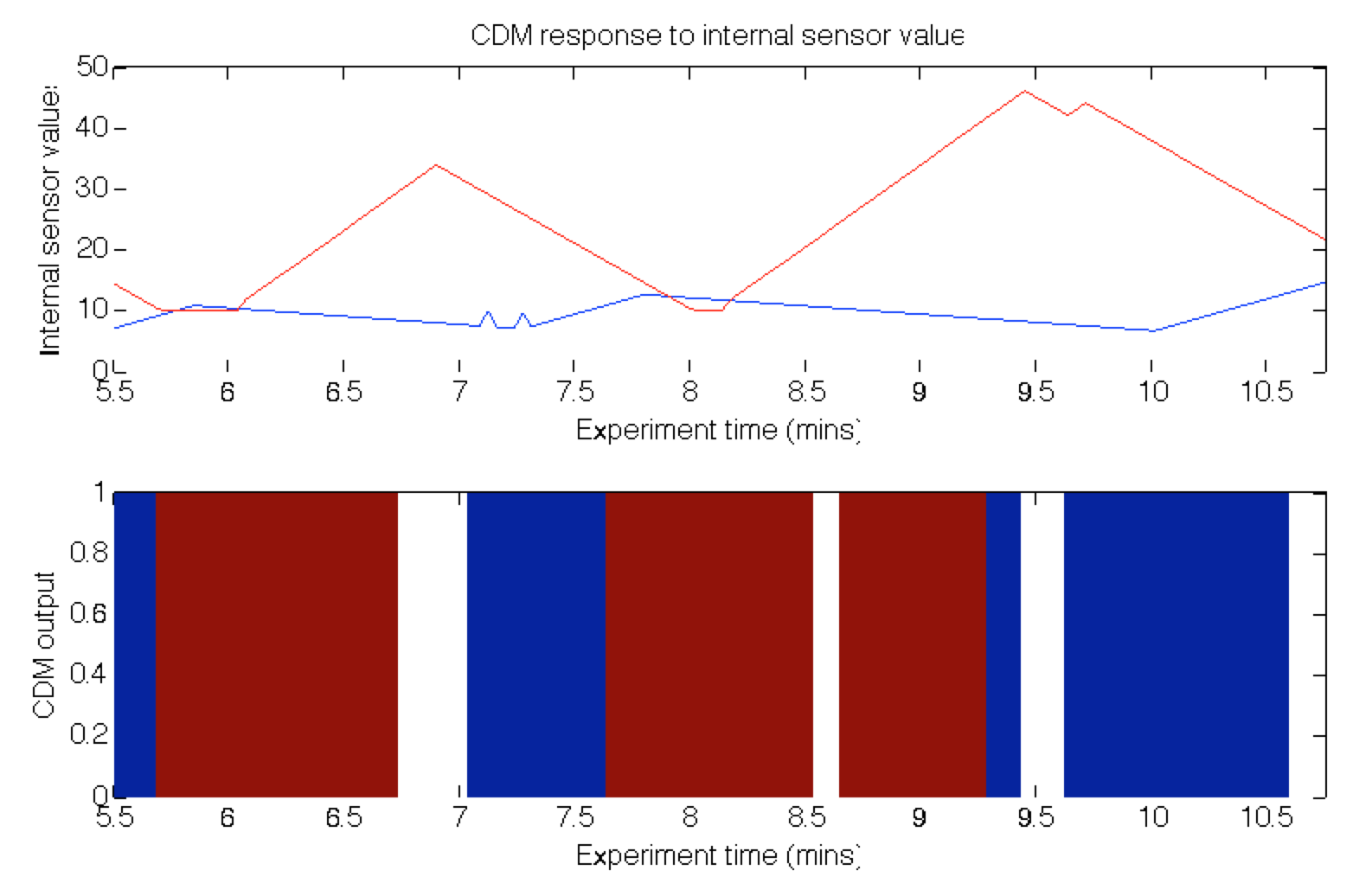} } \caption[Robot behaviour in the arena, along with 
 internal values]{A short (5 min) excerpt from one of the 29-minute replicates. The trace (a) shows how the robot 
 moves between the red and blue areas in the environment as it needs to charge. The graphs in (b) show that as the 
 internal sensor values for charge get too low, the robot actively searches for the charging station, and once it is 
 charged sufficiently, it returns to warming itself under the red lamp.}
 \label{fig:robot:results:example_graphs}
\end{figure}

 Fig.~\ref{fig:robot:results:example_graphs} shows the path of one run (a) with the \cdm{} in place. The graph in (b) 
shows a subset of the internal sensor data provided to the \cdm{} component for the duration of the experimental run. 
The robot was allowed to continue running until it reached 29 minutes, at which point it would be stopped\footnote{the 
limit of 29 minutes is a consequence of the maximum length of time for which the camera could record}. The traces show that 
the robot is able to perform a task while preventing itself from running low on charge.

Table \ref{tab:robot:results:two-sensor} shows the length of time that the robot survived when set up with the two 
components described above (up to the maximum of 29 minutes). This data is presented in a table format, rather than a 
boxplot due to the relatively few replicates in this experiment. The high variance in the \cdm{} data is unexpected 
and can be explained by the robot getting stuck in a corner early on in the run and failing as a result. These 
`failed' runs are still included to ensure the results are not biased by their removal. Even with only 6 replicates, 
it is clear that the \cdm{} code can survive for longer than the control code. This is confirmed through the 
Mann--Whitney--Wilcoxon test and Vargha--Delaney A-test, giving $p<0.05$ and $A>0.84$ respectively. From this, $H_0$ 
can be rejected at the 95\% confidence level.

\begin{table}[h!]
 \centering
 \begin{tabular}{r|l|l}
  \textbf{Replicate}    &       \textbf{Control (mins)}    &       \textbf{Test~(mins)} \\ \hline\hline
  1                             &       3.5     &       29              \\ \hline
  2                             &       4       &       2.5             \\ \hline
  3                             &       2       &       29                      \\ \hline
  4                             &       2.5     &       6                       \\ \hline
  5                             &       10      &       29                      \\ \hline
  6                             &       3       &       29
 \end{tabular}
 \caption[Results from the dual-attractor experimental setup]{Details of how long the
robot survived in the environment when using the \cdm{} (test) component and the threshold (control)
component. Mann--Whitney--Wilcoxon test and Vargha--Delaney A-test give $p<0.05$ and $A>0.84$
respectively, rejecting $H_0$ with 95\% confidence.}
 \label{tab:robot:results:two-sensor}
\end{table}

While performing a relatively simple task, the results presented in this section show that the \cah{}
architecture is a viable option for providing simple artificial homeostasis to a robot. This provides
the evidence required to answer the research question \resQu{2}. The results here also suggest that the
robot is able to make use of its previous experience in the environment, and apply it to survive in the
same environment. This provides further evidence towards answering research question \resQu{1}. The
next section discusses how well the architecture handles more complicated scenarios that involve
conflicting needs.

\subsection{Conflicting decisions}
\label{sec:robot:results:conflict}

While the work above shows how the \cah{} system provides basic homeostatic behaviour to a robot, this section 
addresses research question \resQu{3}: can the \cah{}-controlled robot survive when faced with conflicting decisions 
about what to do?

\begin{figure}[h!]
 \centering
 \includegraphics[width=0.7\linewidth]{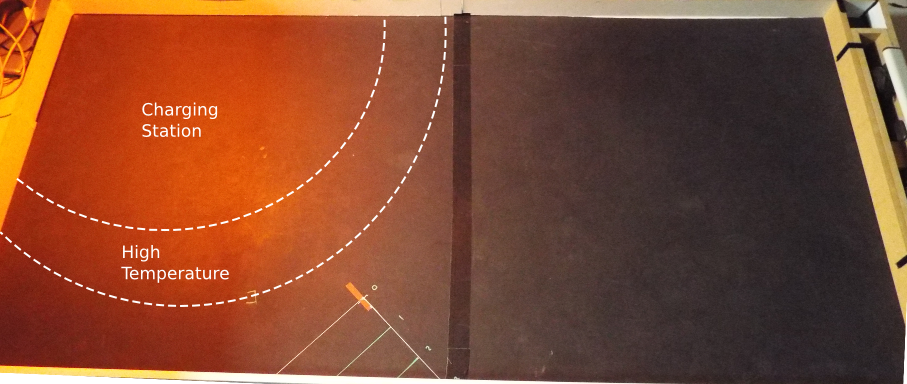}
 \caption[Arena for conflicting decisions]{Photo showing the new arena for conflicting decisions experiment, arena  size: 
224x108cm}
 \label{fig:robot:results:conflict:arena}
\end{figure}

Our experimental setup is altered to bring together the charging station and the higher temperature area (see 
fig.~\ref{fig:robot:results:conflict:arena}). The behaviour of the \cdm{} is inverted with respect to temperature, so 
that the robot now tries to avoid getting warm. In order to charge its battery, therefore, the robot will have to 
endure warmer temperatures. Once the battery runs out, the robot will fail, and once the temperature reaches a set 
maximum, the robot will fail. This experiment once again makes use of the same architecture between test and control 
cases, except for the \cdm{} component that will be replaced by the threshold component for the control case. As 
before, we use a simulated setup initially, and then verify the results using real robot tests.

\nhypothesis A robot running the \cah{} architecture will survive no longer with the \cdm{} component than with the 
threshold component in the presence of conflicting needs. \newline

The experimental setup is as shown in fig.~\ref{fig:robot:results:conflict:arena}. This is the same arena as above, 
except for the removal of the blue light from the right-hand side. In this more difficult scenario, the robot is 
expected to fail more often than it did in the previous setup, as before it was possible for the robot to sit and 
charge indefinitely without failing, whereas here this would result in a failure. After preliminary testing, it was 
evident that the parameter values used previously (given in table~\ref{tab:robot:internal_sensors}) provided a 
scenario that was too difficult for the robot to complete with either \cdm{} or threshold-based architecture (all 
survival times were under 2.5 minutes). As such, the parameter values are adjusted to those shown in 
table~\ref{tab:robot:conflict:internal_sensors}, to make the robot less likely to overheat straight away, but 
balancing this by making it slower to charge.

\begin{table}[h!]
 \centering
 \begin{tabular}{r|c|c}
	~	 		& \textbf{Temperature}		& \textbf{Charge} 	\\ \hline\hline
	{Start value} 		& 10.0				& 5.0			\\ \hline
	{+ve Delta} 		& 1.3				& 0.3			\\ \hline
	{-ve Delta} 		& 0.8				& 0.1			\\ \hline
	{Fail Point}		& 55.0				& 0.1			\\ \hline
	{Limits} 		& [10.0, 60.0]			& [0.01, 15.0]	 	\\ \hline
	{Control threshold}	& 25.0				& 2.5
 \end{tabular} 
 \caption[Parameters for conflicting decision experiment]{The parameters used in the `conflicting decisions' 
 experiment. These values have been picked to result in a challenging, but possible scenario. }
  \label{tab:robot:conflict:internal_sensors}
\end{table}

\begin{figure}[h!]
 \centering

 \subfloat[Simulation environment with white arrow representing the robot and red light representing both the charging 
 station and high temperature area. \label{fig:robot:conflict:simulation:setup}] %
  {\includegraphics[width=0.5\linewidth, valign=c]{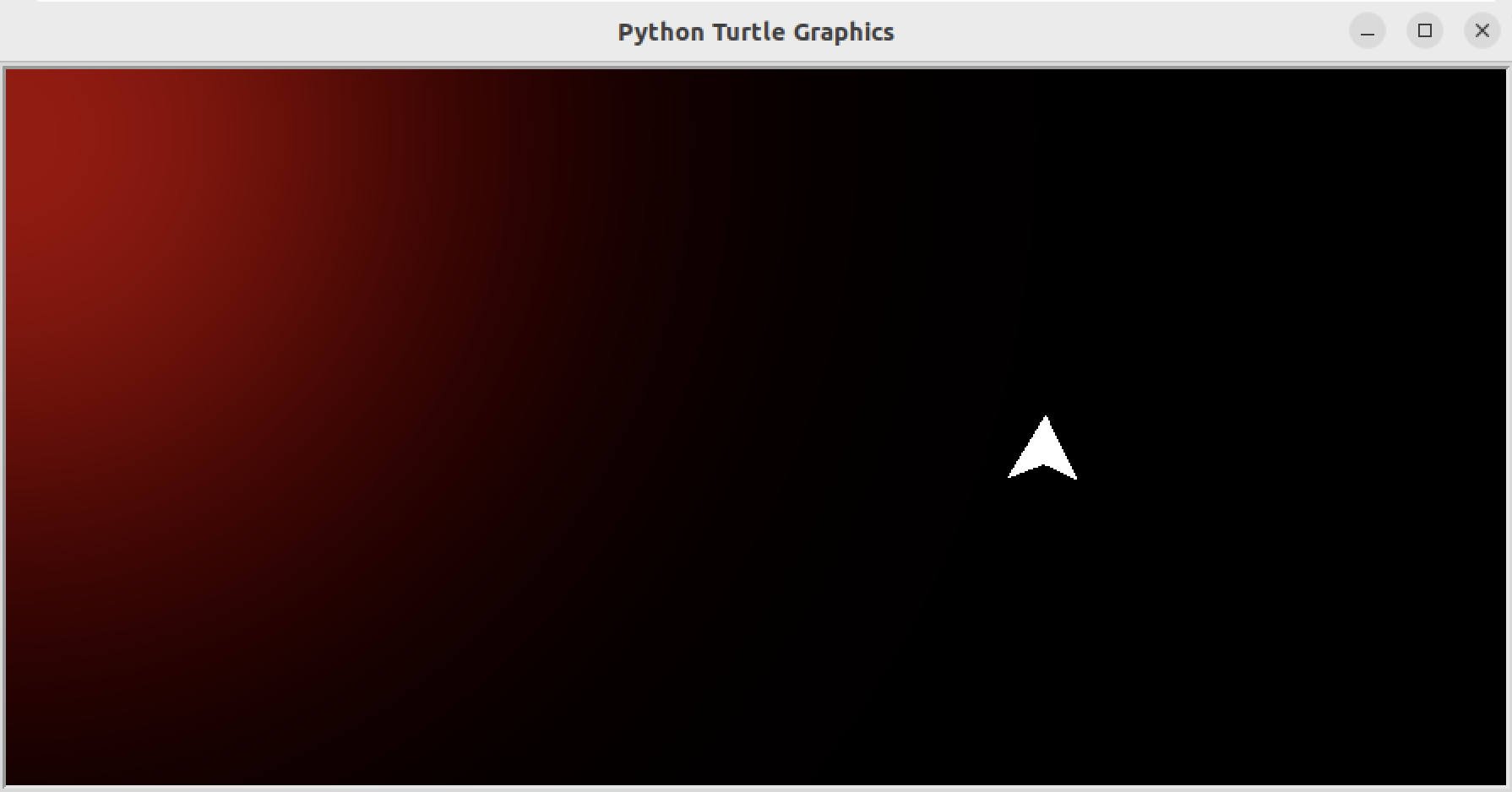} %
   \vphantom{\includegraphics[width=0.4\linewidth,valign=c]{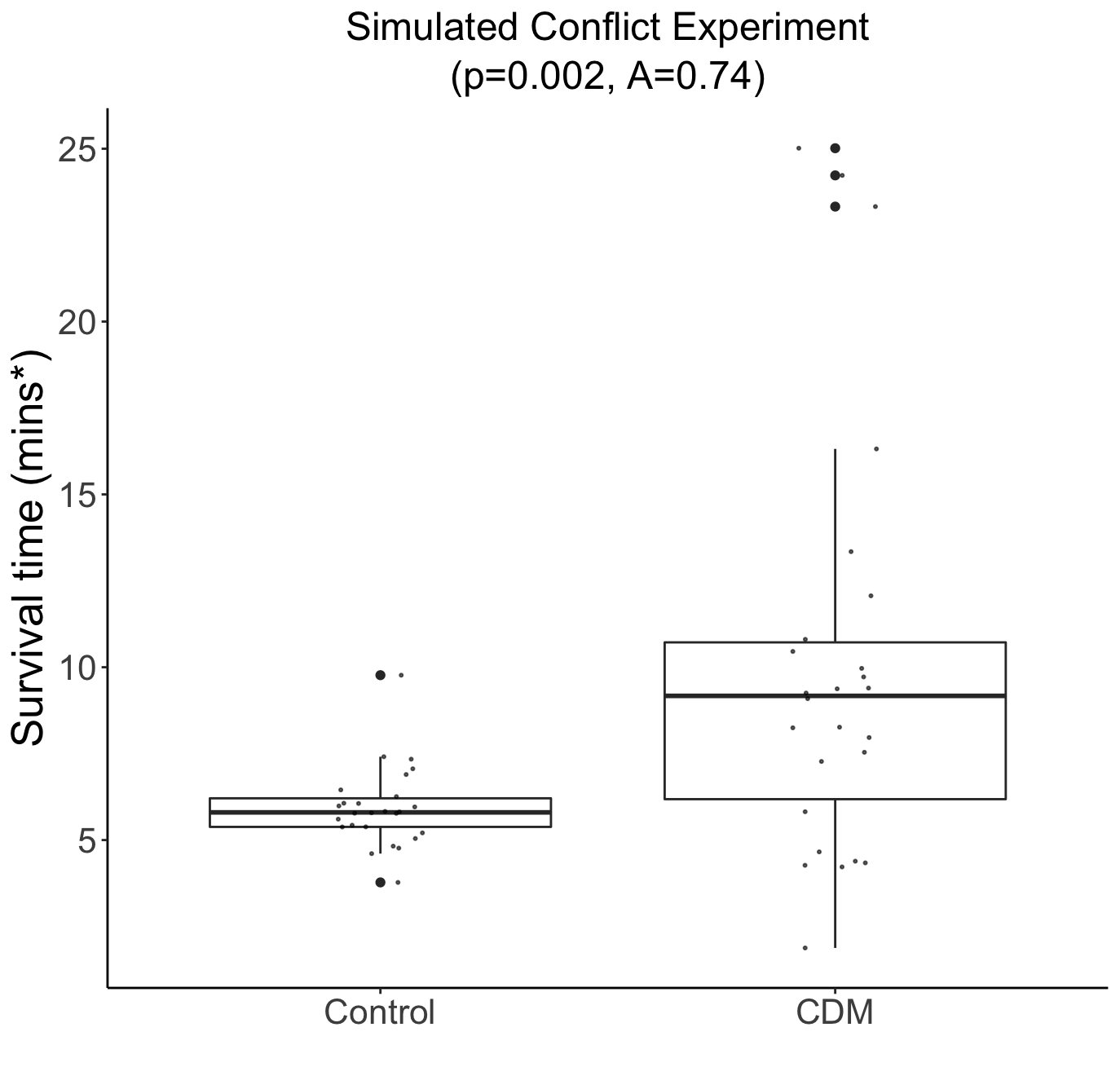}} } \quad
 \subfloat[Boxplots showing that the CDM enables the robot to survive for significantly longer than a simple threshold 
 controller when presented with conflicting decisions (higher is better). The Mann--Whitney--Wilcoxon test and 
 Vargha--Delaney A-test give $p = 0.002$ and $A = 0.74$ respectively, rejecting $H_0$ in simulation with 95\% 
 confidence. \label{fig:robot:conflict:simulation:boxplots}]{\includegraphics[width=0.4\linewidth,valign=c]{sim_conflict.png}} 

 \caption[]{Environment setup and results from simulated experiment (mimicking the real-world setup). N.B. the 
 survival time is measured in minutes, but the simulation runs faster than the real robot, so the results are only 
 proportionally comparable to the results gathered from real robots (presented below).}

 \label{fig:robot:conflict:simulation}
\end{figure}

As with the experiment presented in section~\ref{sec:robot:results:two_sensor}, we use a simulation to demonstrate the 
system behaves as expected without the need for long robot lab experiments. 
Fig.~\ref{fig:robot:conflict:simulation:setup} shows the simulated environment, and 
fig.~\ref{fig:robot:conflict:simulation:boxplots} shows the results of the test. It is clear from 
fig.~\ref{fig:robot:conflict:simulation:boxplots} that the CDM outperforms the control case, as supported by the 
statistical tests showing a significant difference between the distributions.

\begin{figure}[h!]
 \centering

 \subfloat[Trace of robot path between 14 and 19 mins superimposed on the environment. The blue and red circles 
 indicate points where the robot stopped to charge or cool itself respectively. The green and white circles indicate 
 the start and end points. The blue and red solid lines indicate that the robot is searching for, or fleeing from, the 
 red light, respectively. The white dotted line is when the robot wanders (not actively searching for anything).  
 \label{fig:robot:results:conflict:example_graphs:trace} ] { \centering %
 \includegraphics[width=.9\linewidth]{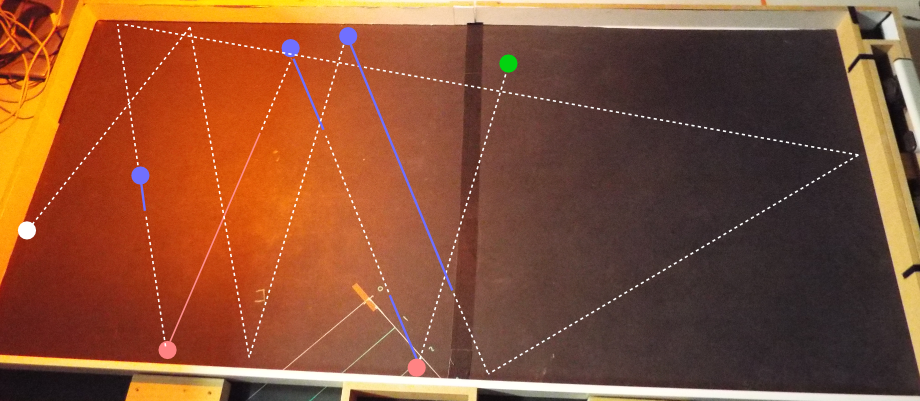} } \\

 \subfloat[Graphs showing internal sensor values (top) and \cdm{} outputs (bottom) for the robot between 14 and 19 
 mins. The red and blue lines show the internal values for temperature and charge, as presented to the \cdm{} 
 component. The red and blue blocks of colour in the bottom graph show the output of the \cdm{} component (red 
 indicating to the robot to flee red light, and blue indicating to search for red light).  
 \label{fig:robot:results:conflict:example_graphs:graphs} ] { \centering %
 \includegraphics[width=.9\linewidth]{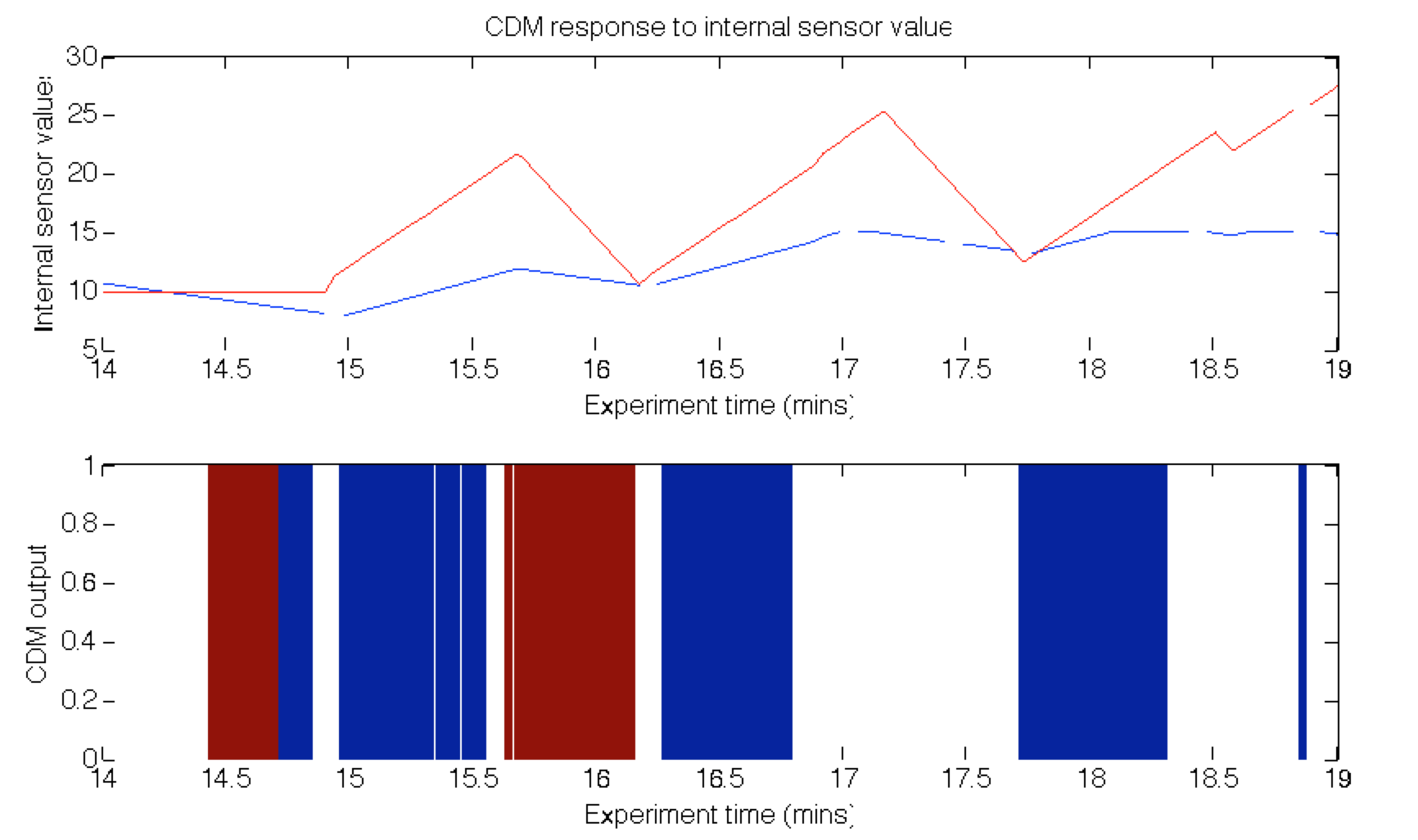} }

 \caption[Robot path trace and graphs from conflict run]{A short (5 min) excerpt from one of the 29-minute replicates. 
 The trace (a) shows how the robot moves in and out of the red area in the environment as it needs to charge and cool 
 down. The graphs in (b) show that as the internal sensor values for charge get too low, the robot actively searches 
 for the charging station, and vice-versa for when the temperature gets too high. Note that, contrary to the behaviour 
 described in fig.~\ref{fig:robot:results:example_graphs}, when the temperature gets too high at around 17 mins, the 
 \cdm{} does not actively push the robot away from the red light, as it is already out of the area by chance. This has 
 the effect of starting to reduce the temperature, reducing the likelihood that the \cdm{} component would push the 
 robot to seek out a cooler area.}

 \label{fig:robot:results:conflict:example_graphs}
\end{figure}

Fig.~\ref{fig:robot:results:conflict:example_graphs} shows a trace of the sensor values as provided to the \cdm{} 
component. The graphs show how the robot balances the need to charge its battery while avoiding getting too warm. The 
robot with \cdm{} component was able to balance the conflicting decisions well, failing only once in 13 replicates, 
whereas the control case failed 6 times out of 13. These results are presented in full in 
table~\ref{tab:robot:results:one_sensor}.

\begin{table}[h!]
 \centering
 \begin{tabular}{r|l|l}
  \textbf{Replicate}    &       \textbf{Control (mins)}    &       \textbf{Test~(mins)} \\ \hline\hline
  1                             &       29     &       29              \\ \hline
  2                             &       3       &       29 	\\ \hline
  3                             &       29       &       29                      \\ \hline
  4                             &       29     &       29                      \\ \hline
  5                             &       9       &       11 	\\ \hline
  6                             &       29       &       29 	\\ \hline
  7                            &       3       &       29 	\\ \hline
  8                             &       29      &       29                      \\ \hline
  9                             &       29       &       29             \\ \hline
  10                             &       8       &       29 	\\ \hline
  11                            &       29       &       29 	\\ \hline
  12                            &       19       &       29 	\\ \hline
  13                            &       20       &       29 	
 \end{tabular}
 \caption[Results from the conflict experimental setup]{Details of how long the
robot survived in the environment when using the \cdm{} (test) component and the threshold (control)
component. Mann--Whitney--Wilcoxon test and Vargha--Delaney A-test give $p=0.03$ and $A=0.70$
respectively, rejecting $H_0$ with 95\% confidence.}
 \label{tab:robot:results:one_sensor}
\end{table}

Table~\ref{tab:robot:results:one_sensor} show the amount of time that the two setups survived (up to the maximum of 29 
minutes). There are 13 replicates for each. While these results were expected to be less clear-cut than the previous 
experiment, analysis gives similar results. The Mann--Whitney--Wilcoxon test gives $p=0.03$, and the Vargha--Delaney 
effect-magnitude test gives $A=0.70$. From this, $H_0$ can be rejected with $95\%$ confidence, resulting in the 
conclusion that the \cdm{} component is better at balancing conflicting decisions than the threshold component, and 
providing sufficient evidence to answer research question \resQu{3}. As with the previous experiment, the results here 
suggest that the robot is still able to make use of its previous experience in the environment, and apply it to 
survive in the same environment. This provides even further evidence towards answering research question \resQu{1}.

\section{Discussion and Future Work}
\label{sec:discussion}

This paper has presented a new approach to developing robotic homeostasis. By taking inspiration from cognition, rather
than cellular-level biology, a simple homeostatic architecture has been developed for a robot which is able to balance
conflicting needs. Evidence towards answering three research questions is presented. These research questions are
summarised below. \newline

\noindent\textbf{\resQu{1}} ({\em Can \cah{} enable a robot learn to from previous experiences and use them to 
influence future behaviour?}) Section \ref{sec:artf_cogn:so:impl} presents the testing of the \ii{} training. This 
process consists of the robot roaming around an environment and associating simultaneous changes in the internal and 
external sensors. These associated changes represent the previous experiences that the robot has learnt. The behaviour 
of the robot in sections \ref{sec:robot:results:two_sensor} and \ref{sec:robot:results:conflict} show that the robot 
is able to successfully recall these experiences and use them to influence its high-level behaviour. \newline

\noindent\textbf{\resQu{2}} ({\em Can \cah{} provide homeostatic behaviour to a robot?}) Section 
\ref{sec:robot:results:two_sensor} presents results showing that the robot is able to alter its high-level behaviour 
according to the value of two internal sensors (temperature and charge). The behaviour of the robot is compared 
between two versions of the \cah{} architecture: one with the \cdm{} component, one with a simple threshold function 
(see section \ref{sec:artf_cogn}). The \cdm{} component consistently outperforms the threshold function at 
providing homeostatic behaviour to a robot, as shown by the survival task in 
fig.~\ref{fig:robot:results:example_graphs}. \newline

\noindent\textbf{\resQu{3}} ({\em Can \cah{} balance two conflicting needs to provide homeostatic behaviour to a 
robot?}) Section \ref{sec:robot:results:conflict} presents results showing that the robot is able to balance two 
conflicting needs. The robot was required to withstand a high-temperature region in order to gain access to a charging 
station. The behaviour of the robot is again compared between two versions of the \cah{} architecture as with 
\resQu{2}, above. The \cdm{} component once again consistently outperforms the threshold function at providing 
homeostatic behaviour to a robot, while balancing two conflicting needs. \newline

%~\citep[\citeyear{palm_towardstheorycellassemblies}]{palm_onassociativememory}  
%neal_timidityusefulemotional, neal_oncemoreunto, 

While other approaches to building artificial homeostasis have directly mimicked the natural systems of the 
body~\citep[\citeauthor{neal_oncemoreunto}, \citeyear{neal_timidityusefulemotional}; 
\citeyear{neal_oncemoreunto}]{vargas_artificialhomeostaticsystem, stradner_analysisimplementationartificial, 
schmickl_modellinghormoneinspired}, the \cah{} architecture has been developed by taking inspiration from 
cognition~\citep{cohen_tendingadamsgarden, mitchell_selfawarenesscontrol}. Combining the \cdm{} (\cdmLongLower{}) with 
a CMM-based internal image module provides the \cah{} (\cahLongUpper{}) architecture with the ability to alter its 
high-level behaviour according to the low-level state of the robot, while based on previous experiences.

The CMM is trained by coincident `spikes' in the internal and external sensors (\eg when the `charge' internal sensor 
rises at the same time as the `blue' external sensor---this relates to a charging station under a blue light in the 
real-world environment). This provides adaptivity to the robot, as it is able to learn about new parts of an 
environment and adapt its behaviour accordingly (for example, if the robot were to find another charging station, it 
would store the association in the CMM and recall it as before).

The robot is able to make decisions about its current internal state through the \cdm{} and transfer this to 
high-level behavioural changes, showing that the \cah{} system provides the capacity for homeostasis to the robot. 
Furthermore, the use of the CMM to recall previous experiences allows the robot to proactively search for, or flee 
from, specific areas in the environment. This proactivity allows for behaviour that is more realistic for real-world 
applications.

The modular nature of the \cah{} architecture, in that the \cdm{} can be easily replaced by a different 
decision-making mechanism, makes the \cah{} architecture more flexible for different applications. It could be 
possible to run multiple decision-making algorithms as an ensemble, with the combination of their outputs being passed 
to the CMM. Furthermore, there is a real possibility to extend this system to alter the perception of a robot based on 
the internal needs of the system. This would open up an interesting line of research where robots could be trained to 
focus their attention only on those parts of their immediate environment that relate to their current task, reducing 
the computational overhead required to solve tasks.

There is also potential for online learning about an environment, where the CMM would be training and recalling 
simultaneously. Furthermore, the sharing of experience between multiple robots working in the same area could be 
possible by sharing the contents of CMMs between robots.

In summary, cognition is a viable source of inspiration for homeostasis in robots. The resulting architecture shows 
that there is significant potential from implementing cognitive behaviour into robots, giving the ability to balance 
conflicting needs more effectively than a conventional threshold mechanism.

%This work represents a first step towards realising this view of altered perceptions. The \cah{} architecture we 
%present in this paper provides the ground work for more advanced cognitive experiments.

\appendix

\section{Experimental Setup}
\label{app:setup}

 The arena is 224x108cm with 15cm high boundaries (see the background of 
fig.~\ref{fig:robot:cdm:arena}). It has coloured LED lamps that can be moved so they point at 
different parts of the arena.

%\begin{figure}[h!]
% \centering
% \includegraphics[width=\linewidth]{arena.jpg}
% \caption[Photo of the laboratory arena]{Photo showing the laboratory arena. The arena is 224x108cm with 15cm high
% boundaries and coloured LED lamps (out of view) that provide environmental cues to the robot.}
% \label{fig:robot:implementation:setup:arena}
%\end{figure}

 The robot platform is the Pi-Swarm \citep{hilder_piswarm}. This platform provides basic IR distance sensing and 
 wheeled movement, along with colour LEDs on the top edge of the robot, and a simple speaker built-in. The robot is 
 controlled using an ARM mbed LPC1768 chip~\citep{arm_mbed} which runs custom C++ code. It has very limited flash and 
 RAM storage, along with limited computational capacity. In order to provide information to the robot about the 
 environment, the Pi-Swarm platform has a single RGB colour sensor (TCS3472) mounted on top.

 LED lamps, mounted on tripods, provide a gradient of colour for the robot, using Perspex colour filters over the 
 front in order to separate the different gradients. After initial analysis the best colours for this laboratory setup 
 (in terms of other lights in the surrounding area and the contrast in the sensor) is a combination of the orange and 
 yellow-green filters (referred to as `red' throughout the paper for simplicity), or the blue filter.

 For the purposes of this work, the robot makes use of two internal sensors: battery charge and temperature. These 
 sensors are simulated, exploiting the fact that they are always beneath a lamp. As such, by setting the threshold for 
 `detecting' a charging station higher than the threshold for detecting the corresponding colour, the charging station 
 is placed at the centre of the lamp's gradient. The colour sensor returns values that are scaled according to the 
 integration time and gain of the sensor (see \citep{tcs3472} for full details). The integration time and gain used 
 for this experimental setup are 14.4ms and 60x, respectively. The colour sensor returns values based on the 
 irradiance of its sensor, which is dependent on a wide range of environmental factors. The values used are 
 proportional to lux, but it makes more sense to report values in terms of the relative responsivity of the sensor. 
 For this reason, we define the `arbitrary colour unit' (acu) as the value returned from the TCS3472 given an 
 integration time of 14.4ms and gain of 60.

\begin{figure}[h!]
 \centering
 \includegraphics[width=0.25\linewidth]{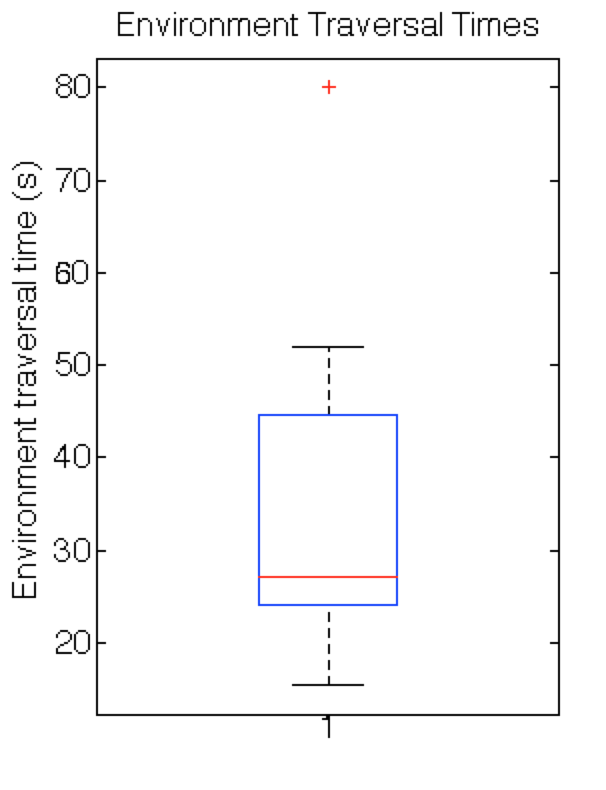}
 \caption[Arena traversal times]{Boxplot showing how long it takes to cross the arena between light
 sources, measured empirically.}
 \label{fig:robot:results:attractor_timing}
\end{figure}

  The threshold value for the control case was calculated from the amount of time it takes for the robot to get to the 
light source it is searching for. This was determined empirically as 45 seconds (the upper quartile of the data 
represented by fig.~\ref{fig:robot:results:attractor_timing}). Using this value for traversing the environment, the 
threshold can be calculated as 2.0. In order to take into account the variability of working in a noisy environment, 
the threshold was increased slightly to 2.5, allowing 54 seconds in the worst case for the robot to find its way 
between light sources. This also helps to reduce the chance of a false positive result by making it easier for the 
control setup to survive for longer.

 \begin{acks}
 Parts of the work presented in this article was undertaken during JHS' PhD~\citep{stovold_thesis}, funded by an 
EPSRC Doctoral Training Grant while at the University of York. A copy of JHS' thesis is available through the White 
Rose eTheses Online portal. 
\end{acks}

\begin{dci}
The authors declare that there is no conflict of interest. 
\end{dci}

%\bibliographystyle{mslapa}
%\bibliographystyle{apa}
%\bibliography{sage}

\end{document}